\documentclass{article} 
\usepackage{iclr2027_conference,times}

\usepackage{amsmath,amsfonts,bm}

\def\eqref#1{equation~\ref{#1}}

\def\1{\bm{1}}

\DeclareMathAlphabet{\mathsfit}{\encodingdefault}{\sfdefault}{m}{sl}
\SetMathAlphabet{\mathsfit}{bold}{\encodingdefault}{\sfdefault}{bx}{n}

\usepackage{graphicx}
\usepackage{hyperref}
\usepackage{url}
\usepackage{booktabs}
\usepackage{multirow}
\usepackage[table]{xcolor}
\usepackage{pgfplots}
\pgfplotsset{compat=1.18}
\usepgfplotslibrary{groupplots,fillbetween}
\usetikzlibrary{positioning,fit,matrix,arrows.meta,backgrounds,calc}
\usepackage{pifont}
\usepackage{wrapfig}
\usepackage{needspace}
\usepackage[skins,breakable]{tcolorbox}

\title{SkillSandbox: Skill Verification via \\Dynamic Scenario Synthesis}

\author{Serin Kim \quad Kwangwook Seo \quad Dokyung Song \quad Jinyoung Yeo \quad Dongha Lee\thanks{Corresponding author.} \\
Yonsei University \\
\texttt{\{kimserin,tommy2130,dokyungs,jinyeo,donalee\}@yonsei.ac.kr}
}

\newcommand{\proposed}{\textsc{SkillSandbox}}

\iclrfinalcopy
\begin{document}

\setcounter{footnote}{1}
\maketitle
\lhead{Preprint}
\addtocontents{toc}{\protect\setcounter{tocdepth}{-1}}

\begin{abstract}
Self-evolving agents distill task-solving experience into skills for future reuse, but these skills can encode incorrect procedures or non-transferable knowledge. 
It is therefore critical to verify each skill’s \emph{reusability}: whether its guidance remains useful beyond the experience from which it was distilled. 
Such verification requires observing how a skill affects execution in new tasks, yet existing tasks may not expose the situations where the target skill can actually be exercised. 
To construct such situations, we propose \proposed, a framework that dynamically synthesizes a task and its environment for each skill that are skill-relevant yet novel. 
A \textsc{Proposer} specifies the conditions to preserve and the source-specific details to vary, a \textsc{Builder} constructs an executable scenario, and a \textsc{Verifier} compares executions with and without the skill. 
The \textsc{Verifier} assesses executability, utility, and efficiency to assign a Keep or Reject verdict, determining whether the skill enters the library. 
Across ALFWorld and WebShop with three models, \proposed{} consistently yields the strongest downstream performance and improved execution efficiency. 
Further analyses examine whether these gains reflect accurate assessment of skill reusability and identify which components of \proposed{} contribute to them.
\href{https://anonymous.4open.science/r/skillsandbox-647C/}{[CODE]}

\end{abstract}

\section{Introduction}
\label{sec:introduction}
\providecommand{\tbd}[1]{\textbf{[#1]}}

Self-evolving agents seek to improve their capabilities by learning from their own experiences~\citep{gao2026surveyselfevolvingagentswhat,yang-etal-2026-towards}.
They distill task-solving experience into skills~\citep{anthropic2025skillsrepo} that provide procedural guidance and store them in a library for future reuse~\citep{wang2023voyager, ouyang2026reasoningbank}.
However, distilled skills do not always guarantee the performance improvement~\citep{li2026skillsbench,han2026swe,gautam2026skillaxe}.
Because task-solving experience contains redundant exploration, mistakes, and task-specific details, distillation may yield incorrect procedures or non-transferable knowledge~\citep{ni2026trace2skill, ma2026skillgen, xiong-etal-2026-memory}.
Once stored and retrieved, such skills can repeatedly interfere with the agent's behavior~\citep{dong2026agent, liang-etal-2026-learning-remember}.
Self-evolving agents therefore need to verify each skill's \emph{reusability}: whether the guidance remains useful when applied beyond the source experience.

Such verification requires execution evidence from a scenario that is both skill-relevant and novel.
That is, the agent should actually encounter the situation the skill addresses in a new environment, so that the evidence reveals whether the guidance improves its behavior beyond the source experience.
Because different skills address different situations, the scenario that provides this evidence is inherently skill-specific.
However, existing approaches~\citep{wang2025inducing,zhang2026agentic,ouyang2026skillos} verify skills against a static pool of tasks that already exists, and merely select from it.
Replaying the source task likely exposes the targeted situation, but only in the context the skill was distilled from, so the evidence is relevant but not novel~\citep{chen2026skillcat,liu2026skillrevise}.
Evaluating the skill on other existing tasks introduces new contexts, but since these tasks are defined independently of the skill, they need not expose the targeted situation.

Verification should therefore shift from \emph{selecting} existing tasks, to \emph{constructing} the task and environment around the target skill, on demand.
To this end, we propose \proposed, a framework that dynamically synthesizes a task and environment where the target situation arises in a context different from the source experience to verify skills.
Each synthesized task and its configured environment form a verification scenario, which serves as a controlled test of that skill.
\proposed{} consists of a \textsc{Proposer}, a \textsc{Builder}, and a \textsc{Verifier}, which design, construct, and evaluate this test, respectively.
(i) The \textsc{Proposer} specifies the conditions that reproduce this situation and the modifications that distinguish the scenario from the source experience.
(ii) The \textsc{Builder} realizes this specification as an executable task and its configured environment.
(iii) The \textsc{Verifier} compares executions in the scenario with and without the skill and measures the effect of the skill in terms of executability, task success, and efficiency.
Consequently, a reusable skill yields a positive effect across its scenarios and is admitted to the library, whereas a skill that is incorrect, overly specific, or not actionable yields no benefit or degrades execution and is rejected.

We evaluate \proposed{} on ALFWorld and WebShop by comparing the library of verified skills with libraries constructed by baseline methods, which either accumulate every generated skill or continually update the library as experience accumulates.
Across three models, the verified library achieves the highest success rate with the fewest steps on both benchmarks.
Further analyses show that the performance gains originate from the synthesized scenarios, as verification on source tasks or randomly selected tasks yields smaller gains and the gap widens on task types unseen in the source experience.
In addition, the verification predicts the helpful and harmful effects of individual skills on held-out tasks.
Each criterion also proves necessary, as the full \textsc{Verifier} outperforms variants that omit executability or efficiency.

Our contributions are as follows:
\begingroup
\setlength{\leftmargini}{2pc}
\begin{itemize}
  \item We identify that skill verification should shift from \emph{selecting} existing tasks to \emph{constructing} the task and environment around the target skill on demand.
  \item We propose \proposed, which synthesizes a task and environment that expose the situation needed to verify the skill. Its \textsc{Proposer} and \textsc{Builder} generate scenarios, and \textsc{Verifier} measures the reusability of skill and determines whether to keep or reject it.
  \item Our experiment confirms that the skill library verified by \proposed{} significantly improves downstream performance. Extensive analyses validate the effect of \proposed{} and show that the synthesized scenarios provide a reliable verification environment.
\end{itemize}
\endgroup

\Needspace*{8\baselineskip}
\suppressfloats[t]
\section{Preliminary Analyses: Skill-Specific Evidence Is Sparse in Existing Experience}
\label{sec:preliminary}
We first ask whether the scenario synthesis is actually necessary to verify skills.
Can existing tasks provide the evidence needed for verification?
To answer this question, we examine alternatives to scenario synthesis and evaluate whether either one reliably provides the skill-specific evidence.
\begingroup
\setlength{\leftmargini}{2pc}
\begin{itemize}
    \item \textbf{Question I:} \emph{Do semantically similar tasks expose the skill-relevant situation?}
    \item \textbf{Question II:} \emph{How sparse are skill-relevant situations in downstream experience?}
\end{itemize}
\endgroup

\subsection{Question I: Do semantically similar tasks expose the skill-relevant situation?}

\begin{wraptable}{r}{0.46\textwidth}
\vspace{-2\baselineskip}
\centering
\small
\setlength{\belowcaptionskip}{4pt}%
\setlength{\tabcolsep}{2pt}
\renewcommand{\arraystretch}{1.12}
\caption{Mean skill-relevant tasks out of 10 retrieved tasks. $N$: candidate task count.}
\label{tab:prelim_applicability}
\begin{tabular*}{\linewidth}{@{\extracolsep{\fill}}lcccc@{}}
\toprule
\textbf{Executor} & \textbf{ALFWorld} & \multicolumn{3}{c}{\textbf{WebShop}}\\
\cmidrule(lr){2-2}\cmidrule(l){3-5}
$N$ & 140 & 200 & 300 & 500\\
\midrule
Gemini 3.1 Flash & 5.96 & 5.96 & 5.90 & 6.10\\
Qwen3.5-27B & 6.00 & 6.60 & 6.58 & 6.66\\
\bottomrule
\end{tabular*}
\vspace{-1.5\baselineskip}
\end{wraptable}

We examine whether semantic similarity can identify existing tasks that expose the skill-relevant situation.
We use 50 skills per benchmark, 140 held-out tasks in ALFWorld~\citep{shridhar2020alfworld} and nested pools of 200, 300, and 500 tasks in WebShop~\citep{yao2022webshop}, keeping the same 50 skills across pool sizes.
For each skill, we select the 10 held-out tasks whose instructions are most similar to the skill text.
The agent executes each selected task with the target skill provided, and an LLM judge determines from the trajectory whether execution encounters the skill-relevant situation.

\textbf{Instruction similarity does not reliably identify tasks that exercise the target skill.}
As shown in Table~\ref{tab:prelim_applicability}, the skill-relevant situation is observed in only up to 6.66 out of 10 selected tasks per skill across the settings.
Even expanding the candidate pool from 200 to 500 tasks yields only marginal gains in skill-relevant coverage among the top-10 retrieved tasks.
The main reason to this is that semantic similarity is measured at the task level, whereas skill relevance depends on what occurs during execution.
Tasks with similar instructions can still induce different states, interactions, and intermediate decisions, and therefore need not expose the same skill-relevant situation.
Consequently, task-level similarity does not reliably provide the execution evidence required for skill verification.

\subsection{Question II: How sparse are skill-relevant situations in downstream experience?}

\input{fig_prelim_stream}
A second alternative is to accumulate verification evidence from subsequent downstream experience~\citep{zhang2026agentic,xiong-etal-2026-memory,ouyang2026skillos}.
We construct a library of 50 skills and stream 500 held-out tasks in a fixed random order, repeating each setting three times.
For each task, referenced skill IDs are recorded in rationale and counted.
We vary the number of skills accessible, with \textbf{Top-3} and \textbf{Top-5 Retrieval} providing the top-3 or 5 skills retrieved per task, respectively, and \textbf{Full Library} all 50 skills on every task.
Figure~\ref{fig:prelim_stream} (a) reports the percentage of skills used in at least one task, and (b) reports the percentage used in at least 5 distinct tasks.

\textbf{Skill-relevant situations remain sparse as downstream experience accumulates.}
(a) shows that coverage grows slowly after 300 tasks. Executing 200 additional tasks raises the percentage of skills used at least once by only 4.7--6.7 percentage points.
After 500 tasks, only 68\% of skills are used under Top-3 Retrieval, 83\% under Top-5, and 70\% under Full Library.
Thus, depending on the access setting, 17--32\% of skills remain \emph{unexercised}, which means that their effects on the agent's behavior are never observed, leaving their reusability unverifiable.
More reliable verification benefits from repeated evidence across distinct tasks, yet such opportunities are even more sparse.
As shown in (b), only 44--55\% of skills are used in at least 5 tasks.
Even after 500 tasks, 45--56\% of the skills fail to accumulate five execution opportunities.
Overall, downstream experience leaves a substantial number of skills with insufficient evidence for verification.

Together, Questions I and II reveal complementary limitations of relying on existing tasks for verification.
Neither reliably provides the skill-specific evidence that verification needs.
We therefore construct verification scenarios that induce the target situation beyond its source experience.

\vspace{-8pt}
\section{Methodology}
\label{sec:methodology}
\providecommand{\tbd}[1]{\textbf{[#1]}}

\vspace{-6pt}%
\subsection{Problem Setup}
\label{sec:setup}

We model an environment as $E=(\mathcal{S},\mathcal{A},\mathcal{O},T,\kappa)$,
where $\mathcal{S}$, $\mathcal{A}$, and $\mathcal{O}$ are the state, action, and observation spaces, $T$ is the state-transition rule, and the configuration $\kappa$ specifies the environment's objects, arrangements, task-related data, and presentation of information.
A task $x$ specifies the goal and requirements.
At each step, an agent $\pi$ receives a partial observation~\citep{Kaelbling1998PlanningAA} $o_t\in\mathcal{O}$ of state $s_t$ and takes an action $a_t\in\mathcal{A}$, yielding $s_{t+1}\sim T(\cdot\mid s_t,a_t)$, and the trajectory is $\tau_x=(o_0,a_0,o_1,a_1,\ldots)$.
We call the environment and information encountered during execution the \emph{task context}.

{\setlength{\abovedisplayskip}{1pt}\setlength{\belowdisplayskip}{1pt}%
A synthesis specification $\delta$ programmatically specifies the requirements and changes that construct a new task $x_{\delta}$ and configuration $\kappa_{\delta}$, which with the success criterion $g_{\delta}$ define
\[
E_{\delta}=(\mathcal{S},\mathcal{A},\mathcal{O},T,\kappa_{\delta}),
\qquad
\sigma_{\delta}=(x_{\delta},E_{\delta},g_{\delta}).
\]
\par}

\subsection{Skill Verification}
\label{sec:verification}
\begin{figure}[t]
\centering
\includegraphics[width=\linewidth]{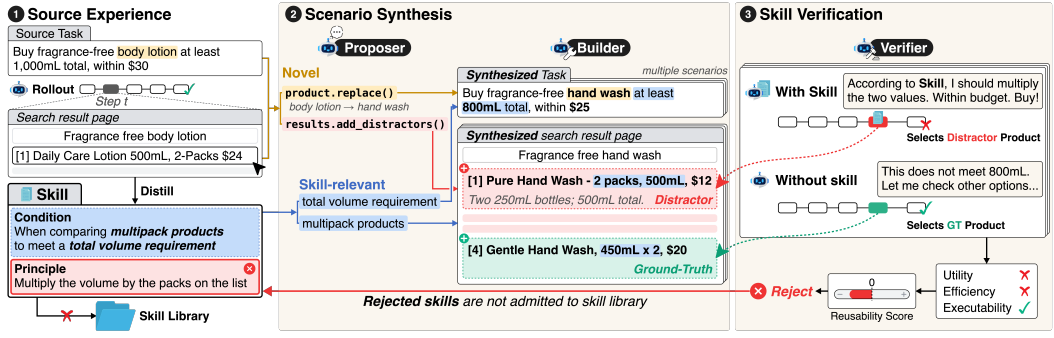}
\caption{Overview of skill verification via dynamic scenario synthesis.}
\label{fig:main}
\end{figure}

A skill $k=(\texttt{condition},\texttt{principle})$ is distilled from a source task $x_{\mathrm{src}}$ and trajectory $\tau_{\mathrm{src}}$.
The \texttt{condition} describes when the skill applies, and the \texttt{principle} provides knowledge or guidance for that situation.
Verification assesses the skill's \emph{reusability}, defined as its ability to guide executable behavior that improves task success or execution efficiency in new situations satisfying the \texttt{condition}.
We verify the skill's reusability along three dimensions:
\begingroup
\setlength{\leftmargini}{2pc}
\begin{enumerate}
  \item \textbf{Executability:} The \texttt{principle} can be executed through admissible actions in the environment when the \texttt{condition} is satisfied.
  \item \textbf{Utility:} Applying the \texttt{principle} helps the agent achieve the task goal, improving task success relative to execution without the skill.
  \item \textbf{Efficiency:} The \texttt{principle} enables the task completion with fewer actions than execution without the skill.
\end{enumerate}
\endgroup

Evaluating these dimensions only on the source task $x_{\mathrm{src}}$ does not establish reusability in new task contexts, while arbitrary tasks may not reproduce the situation described by the \texttt{condition}.
Verification scenarios must therefore be both \emph{skill-relevant} and \emph{novel}.
A scenario is skill-relevant when completing the task requires addressing the situation described by \texttt{condition}, in which the guidance in \texttt{principle} can be evaluated.
It is novel when the agent encounters a task context that differs from the source experience.

\subsection{Scenario Synthesis}
\label{sec:synthesis}
To construct scenarios that are both skill-relevant and novel, we reproduce the situation described by the \texttt{condition} while reconstructing the new task context.
We jointly construct a task $x_{\delta}$ and configuration $\kappa_{\delta}$ so that the \texttt{condition} holds during execution, while varying the details.
Execution in the resulting scenario tests whether the \texttt{principle} remains useful when the situation described by the \texttt{condition} occurs in a context different from the source experience.
Figure~\ref{fig:main} illustrates this synthesis process.
In \proposed, the \textsc{Proposer} specifies how to reproduce the condition in a new context, and the \textsc{Builder} constructs the scenario according to this specification.

\paragraph{\textsc{Proposer}.}
The \textsc{Proposer} first uses the \texttt{condition}, $x_{\mathrm{src}}$, and $\tau_{\mathrm{src}}$ to jointly plan reproduction and reconstruction.
It encodes its natural language plan as a programmatic specification $\delta$, using shared construction operations exposed by the construction interface $\mathcal{I}_E$.
This interface specifies which task and configuration fields can be modified (e.g., \texttt{list\_order}, \texttt{product\_option}) and the operations permitted for each field (e.g., \texttt{add\_distractor}, \texttt{remove}, \texttt{alter}).
The specification includes variables for objects and values that the \textsc{Builder} later instantiates.
For reproduction, it specifies the requirements that the task query and configuration must satisfy for the situation described by the \texttt{condition} to occur during execution.
For reconstruction, it specifies the modifications that produce a context different from the source experience while satisfying the reproduction requirements.
The resulting specification $\delta$ is expressed by the operations permitted by $\mathcal{I}_E$, providing the \textsc{Builder} with a plan for scenario construction.

\paragraph{\textsc{Builder}.}
To construct an executable scenario according to given $\delta$, the \textsc{Builder} uses the task and configuration data $\mathcal{D}_E$ (e.g., \texttt{product\_database} for WebShop, \texttt{object\_list} for AlfWorld) available in $E$.
The \textsc{Builder} first binds the variables in $\delta$ to concrete objects and values from $\mathcal{D}_E$.
It then translates the specified operations into an executable construction program for $E$ and runs it.
As specified in $\delta$, this program constructs new task instruction, environment, and corresponding ground truth, using $\mathcal{D}_E$.

To faithfully reproduce the skill's \texttt{condition} across diverse contexts, this construction proceeds within a validation loop.
Using the same LLM model as the executor, the loop checks whether each constructed scenario satisfies the requirements in the \textsc{Proposer}'s specification $\delta$ and whether it is identical to any previously generated scenario for the same skill.
Scenarios that fail validation are discarded, and new candidates are generated until five valid scenarios are obtained.
Together, $x_{\delta}$ and $E_{\delta}$ reproduce the situation described by the \texttt{condition} in a reconstructed task context, allowing the agent to reuse the \texttt{principle} beyond the source experience.
The \textsc{Builder} returns the resulting verification scenario $\sigma_{\delta}$:
\[
\delta=\textsc{Proposer}(Skill,x_{\mathrm{src}},\tau_{\mathrm{src}},\mathcal{I}_E), \qquad \sigma_{\delta}=\textsc{Builder}_E(\delta;\mathcal{D}_E).
\]
To illustrate what kind of scenarios are generated and how they are constructed, we provide modification statistics (Appendix~\ref{app:scenario_synthesis}), and case studies (Appendix~\ref{app:case_study}).

\paragraph{\textsc{Verifier}.}
For each synthesized scenario, the \textsc{Verifier} compares trajectories $\tau^{+}$ and $\tau^{-}$ produced by the same executor with and without skill $k$ (Figure~\ref{fig:main}, right).
At aligned steps with identical pre-action observations, $\Delta u$ measures the difference in terminal success rewards discounted by the remaining action counts, capturing utility and efficiency.
The executability indicator $e$ is $1$ when the action in $\tau^{+}$ implements the skill's instruction and $0$ otherwise.
We average $e\,\Delta u$ over retained step pairs within each trajectory pair and then across scenarios to score reusability and assign a verdict (score calculation are detailed in Appendix~\ref{app:verifier_details}):
\begin{equation}
R(k)=\widehat{\mathbb{E}}[e\,\Delta u],
\qquad
\textsc{Verifier}(k)=
\begin{cases}
\textsc{Keep}, & R(k)>0,\\
\textsc{Reject}, & R(k)\leq 0
\end{cases}.
\label{eq:skill_retention}
\end{equation}

\section{Experiments}
\label{sec:experiments}
\providecommand{\tbd}[1]{\textbf{[#1]}}

We first present the experimental setup (Section~\ref{sec:exp_setup}) and main results (Section~\ref{sec:main_results}).
We compare \proposed{} with baselines that construct skill libraries with or without verification.
To examine how effectively each method retains reusable skills, we build a library using each method and evaluate it on the same held-out tasks.

\subsection{Experimental Setup}
\label{sec:exp_setup}

\paragraph{Benchmarks.}
We evaluate on ALFWorld~\citep{shridhar2020alfworld} and WebShop~\citep{yao2022webshop}.
ALFWorld provides six household task types, and WebShop requires agents to find and purchase products that satisfy user instructions.

\paragraph{Metrics.}
We report success rate (SR, \%) on both benchmarks, average task score (Score, 0--100) on WebShop, and average steps across all episodes.
To characterize how a skill library changes individual task outcomes beyond SR, we additionally report \textsf{Recovery} (\textsf{Rec.}) and \textsf{Regression} (\textsf{Reg.}) relative to the run without skill library of the same model.
\textsf{Recovery} is the percentage of all held-out tasks that fail without skills but succeed with the library, whereas \textsf{Regression} is the percentage that succeed without skills but fail with the library.

\begin{table*}[t]
\centering
\fontsize{9}{10.5}\selectfont%
\setlength{\tabcolsep}{2.0pt}
\renewcommand{\arraystretch}{1.00}%
\setlength{\aboverulesep}{1.2pt}
\setlength{\belowrulesep}{1.8pt}
\newcommand{\verificationgrouprule}{%
  \noalign{\vskip1pt
    \hbox to \textwidth{\color{black!55}\leaders\hbox{\rule{2pt}{0.35pt}\kern2pt}\hfill}%
    \vskip1pt}}
\caption{Main results on held-out ALFWorld and WebShop tasks.
Vanilla Skill, ReasoningBank, and MemP retain skills without verification; ExpeL uses source-task replay, ACE and SkillOS use downstream feedback, and \proposed{} uses synthesized scenarios for verification.
Bold marks the best value per model; colored percentages are relative changes from No Skill.}
\label{tab:main}
\begin{tabular*}{\textwidth}{@{\extracolsep{\fill}} l cccc ccccc @{}}
\toprule
\multirow{2}{*}{\textbf{Method}} &
\multicolumn{4}{c}{\textbf{ALFWorld}} & \multicolumn{5}{c}{\textbf{WebShop}} \\
\cmidrule(lr){2-5} \cmidrule(lr){6-10}
& SR\kern0.5pt$\uparrow$ & Steps\kern0.5pt$\downarrow$ & \textsf{Rec.}\kern0.5pt$\uparrow$ & \textsf{Reg.}\kern0.5pt$\downarrow$ & SR\kern0.5pt$\uparrow$ & Score\kern0.5pt$\uparrow$ & Steps\kern0.5pt$\downarrow$ & \textsf{Rec.}\kern0.5pt$\uparrow$ & \textsf{Reg.}\kern0.5pt$\downarrow$ \\
\midrule[0.8pt]
\rowcolor{gray!15}[\tabcolsep][0pt]
\multicolumn{10}{@{}l@{}}{\textbf{Qwen3.5-9B}} \\
No Skill & 32.1 & 23.8 & -- & -- & 18.8 & 47.1 & 8.0 & -- & -- \\
\midrule
Vanilla Skill & 57.9{\fontsize{6}{6}\selectfont\textcolor{green!45!black}{+80.4\%}} & 19.3 & 30.0 & 4.3 & 15.6{\fontsize{6}{6}\selectfont\textcolor{red!75!black}{-17.0\%}} & 38.0{\fontsize{6}{6}\selectfont\textcolor{red!75!black}{-19.3\%}} & 9.6 & 5.0 & 8.2 \\
ReasoningBank & 53.6{\fontsize{6}{6}\selectfont\textcolor{green!45!black}{+67.0\%}} & 20.0 & 27.1 & 5.7 & 17.2{\fontsize{6}{6}\selectfont\textcolor{red!75!black}{-8.5\%}} & 38.4{\fontsize{6}{6}\selectfont\textcolor{red!75!black}{-18.6\%}} & 10.0 & 5.8 & 7.4 \\
MemP & 42.9{\fontsize{6}{6}\selectfont\textcolor{green!45!black}{+33.6\%}} & 22.1 & 14.3 & 3.6 & 17.2{\fontsize{6}{6}\selectfont\textcolor{red!75!black}{-8.5\%}} & 41.2{\fontsize{6}{6}\selectfont\textcolor{red!75!black}{-12.6\%}} & 9.4 & 6.6 & 8.2 \\
\verificationgrouprule
ExpeL & 42.1{\fontsize{6}{6}\selectfont\textcolor{green!45!black}{+31.2\%}} & 22.1 & 13.6 & 3.6 & 21.8{\fontsize{6}{6}\selectfont\textcolor{green!45!black}{+16.0\%}} & 52.9{\fontsize{6}{6}\selectfont\textcolor{green!45!black}{+12.4\%}} & 7.0 & \textbf{8.2} & 5.2 \\
\verificationgrouprule
ACE & 45.0{\fontsize{6}{6}\selectfont\textcolor{green!45!black}{+40.2\%}} & 21.7 & 18.6 & 5.7 & 15.6{\fontsize{6}{6}\selectfont\textcolor{red!75!black}{-17.0\%}} & 35.4{\fontsize{6}{6}\selectfont\textcolor{red!75!black}{-24.9\%}} & 10.4 & 5.2 & 8.4 \\
SkillOS & 40.0{\fontsize{6}{6}\selectfont\textcolor{green!45!black}{+24.6\%}} & 22.7 & 14.3 & 6.4 & 11.4{\fontsize{6}{6}\selectfont\textcolor{red!75!black}{-39.4\%}} & 25.2{\fontsize{6}{6}\selectfont\textcolor{red!75!black}{-46.6\%}} & 12.2 & 3.8 & 11.2 \\
\verificationgrouprule
\textbf{\proposed} & \textbf{66.4}{\fontsize{6}{6}\selectfont\textcolor{green!45!black}{+106.9\%}} & \textbf{17.5} & \textbf{36.4} & \textbf{2.1} & \textbf{22.2}{\fontsize{6}{6}\selectfont\textcolor{green!45!black}{+18.1\%}} & \textbf{59.1}{\fontsize{6}{6}\selectfont\textcolor{green!45!black}{+25.5\%}} & \textbf{7.0} & 8.0 & \textbf{4.6} \\
\midrule[0.8pt]
\rowcolor{gray!15}[\tabcolsep][0pt]
\multicolumn{10}{@{}l@{}}{\textbf{Gemini 3.1 Flash-Lite}} \\
No Skill & 56.4 & 19.9 & -- & -- & 25.6 & 58.9 & 5.6 & -- & -- \\
\midrule
Vanilla Skill & 49.3{\fontsize{6}{6}\selectfont\textcolor{red!75!black}{-12.6\%}} & 20.6 & 6.4 & 13.6 & 27.6{\fontsize{6}{6}\selectfont\textcolor{green!45!black}{+7.8\%}} & 54.9{\fontsize{6}{6}\selectfont\textcolor{red!75!black}{-6.8\%}} & 8.0 & 7.2 & 5.2 \\
ReasoningBank & 52.9{\fontsize{6}{6}\selectfont\textcolor{red!75!black}{-6.2\%}} & 20.3 & 9.3 & 12.9 & 23.4{\fontsize{6}{6}\selectfont\textcolor{red!75!black}{-8.6\%}} & 53.7{\fontsize{6}{6}\selectfont\textcolor{red!75!black}{-8.9\%}} & 6.9 & 5.4 & 7.6 \\
MemP & 55.0{\fontsize{6}{6}\selectfont\textcolor{red!75!black}{-2.5\%}} & 19.5 & \textbf{12.1} & 13.6 & 25.0{\fontsize{6}{6}\selectfont\textcolor{red!75!black}{-2.3\%}} & 55.3{\fontsize{6}{6}\selectfont\textcolor{red!75!black}{-6.1\%}} & 6.3 & 5.0 & 5.6 \\
\verificationgrouprule
ExpeL & 50.0{\fontsize{6}{6}\selectfont\textcolor{red!75!black}{-11.3\%}} & 21.3 & 9.3 & 15.7 & 27.4{\fontsize{6}{6}\selectfont\textcolor{green!45!black}{+7.0\%}} & 59.4{\fontsize{6}{6}\selectfont\textcolor{green!45!black}{+0.8\%}} & 5.8 & 6.0 & 4.2 \\
\verificationgrouprule
ACE & 42.9{\fontsize{6}{6}\selectfont\textcolor{red!75!black}{-23.9\%}} & 22.0 & 5.7 & 19.3 & 20.8{\fontsize{6}{6}\selectfont\textcolor{red!75!black}{-18.8\%}} & 41.5{\fontsize{6}{6}\selectfont\textcolor{red!75!black}{-29.5\%}} & 10.4 & 6.6 & 11.4 \\
SkillOS & 51.4{\fontsize{6}{6}\selectfont\textcolor{red!75!black}{-8.9\%}} & 20.8 & 10.7 & 15.7 & 28.4{\fontsize{6}{6}\selectfont\textcolor{green!45!black}{+10.9\%}} & 56.4{\fontsize{6}{6}\selectfont\textcolor{red!75!black}{-4.3\%}} & 7.2 & 7.4 & 4.6 \\
\verificationgrouprule
\textbf{\proposed} & \textbf{60.0}{\fontsize{6}{6}\selectfont\textcolor{green!45!black}{+6.4\%}} & \textbf{19.1} & \textbf{12.1} & \textbf{8.6} & \textbf{36.2}{\fontsize{6}{6}\selectfont\textcolor{green!45!black}{+41.4\%}} & \textbf{64.8}{\fontsize{6}{6}\selectfont\textcolor{green!45!black}{+10.0\%}} & \textbf{5.3} & \textbf{12.4} & \textbf{1.8} \\
\midrule[0.8pt]
\rowcolor{gray!15}[\tabcolsep][0pt]
\multicolumn{10}{@{}l@{}}{\textbf{Qwen3.5-27B}} \\
No Skill & 75.7 & 15.4 & -- & -- & 32.2 & 55.4 & 8.0 & -- & -- \\
\midrule
Vanilla Skill & 75.0{\fontsize{6}{6}\selectfont\textcolor{red!75!black}{-0.9\%}} & 15.0 & 7.9 & \textbf{8.6} & 31.8{\fontsize{6}{6}\selectfont\textcolor{red!75!black}{-1.2\%}} & 54.5{\fontsize{6}{6}\selectfont\textcolor{red!75!black}{-1.7\%}} & 8.5 & 8.8 & 9.2 \\
ReasoningBank & 67.1{\fontsize{6}{6}\selectfont\textcolor{red!75!black}{-11.4\%}} & 17.1 & 10.7 & 19.3 & 29.2{\fontsize{6}{6}\selectfont\textcolor{red!75!black}{-9.3\%}} & 53.0{\fontsize{6}{6}\selectfont\textcolor{red!75!black}{-4.4\%}} & 8.4 & 5.4 & 8.4 \\
MemP & 52.9{\fontsize{6}{6}\selectfont\textcolor{red!75!black}{-30.1\%}} & 19.8 & 2.9 & 25.7 & 32.4{\fontsize{6}{6}\selectfont\textcolor{green!45!black}{+0.6\%}} & 57.5{\fontsize{6}{6}\selectfont\textcolor{green!45!black}{+3.8\%}} & 8.6 & 7.2 & 7.0 \\
\verificationgrouprule
ExpeL & 70.0{\fontsize{6}{6}\selectfont\textcolor{red!75!black}{-7.5\%}} & 16.2 & 10.0 & 15.7 & 34.6{\fontsize{6}{6}\selectfont\textcolor{green!45!black}{+7.5\%}} & 54.7{\fontsize{6}{6}\selectfont\textcolor{red!75!black}{-1.2\%}} & 8.3 & 9.6 & 7.2 \\
\verificationgrouprule
ACE & 74.3{\fontsize{6}{6}\selectfont\textcolor{red!75!black}{-1.8\%}} & 15.5 & 12.1 & 13.6 & 37.4{\fontsize{6}{6}\selectfont\textcolor{green!45!black}{+16.1\%}} & 59.7{\fontsize{6}{6}\selectfont\textcolor{green!45!black}{+7.9\%}} & 7.4 & 11.4 & 6.2 \\
SkillOS & 63.6{\fontsize{6}{6}\selectfont\textcolor{red!75!black}{-16.0\%}} & 18.1 & 10.7 & 22.9 & 30.4{\fontsize{6}{6}\selectfont\textcolor{red!75!black}{-5.6\%}} & 49.7{\fontsize{6}{6}\selectfont\textcolor{red!75!black}{-10.3\%}} & 9.0 & 8.0 & 9.8 \\
\verificationgrouprule
\textbf{\proposed} & \textbf{80.0}{\fontsize{6}{6}\selectfont\textcolor{green!45!black}{+5.7\%}} & \textbf{14.3} & \textbf{12.9} & \textbf{8.6} & \textbf{45.0}{\fontsize{6}{6}\selectfont\textcolor{green!45!black}{+39.8\%}} & \textbf{62.2}{\fontsize{6}{6}\selectfont\textcolor{green!45!black}{+12.2\%}} & \textbf{7.2} & \textbf{18.0} & \textbf{5.2} \\
\bottomrule
\end{tabular*}
\end{table*}
\paragraph{Baselines.}
\emph{No Skill} executes tasks without a skill library.
\emph{Vanilla Skill} retains all skills generated with the distillation prompt from SkillRL~\citep{xia2026skillrl}.
We additionally compare with ReasoningBank~\citep{ouyang2026reasoningbank} and MemP~\citep{fang2026memp}, as well as execution-based library curation methods including ExpeL~\citep{zhao2024expel}, ACE~\citep{zhang2026agentic}, and SkillOS~\citep{ouyang2026skillos}.
ExpeL, ACE, and SkillOS admit skills using execution feedback: ExpeL extracts insights from repeated attempts at the source task, whereas ACE and SkillOS update the library from subsequent tasks that are selected independently of any skill.
Detailed descriptions of each baseline are provided in Appendix~\ref{app:baselines}.
\proposed{} retains only the skills from Vanilla Skill that are verified as Keep.

\paragraph{Implementation details.}
We use Qwen3.5-9B~\citep{qwen2026qwen35}, Gemini 3.1 Flash-Lite~\citep{google2026gemini31flashlite}, and Qwen3.5-27B~\citep{qwen2026qwen35} as executors within the ReAct framework~\citep{yao2022react}.
The same model is used for execution and skill generation, and our scenario synthesis and verification also use the same model.
All baselines use the same 100 source tasks to execute and generate skills, and each library is frozen before held-out evaluation.
We evaluate on 140 tasks from the official ALFWorld test sets and 500 tasks from the official WebShop test sets, retrieving the top-3 skills for each task (Appendix~\ref{app:results}).

\subsection{Main Results}
\label{sec:main_results}

\paragraph{Unverified skills can degrade agent performance.}
As shown in Table~\ref{tab:main}, libraries admitted without our verification achieve lower SR than No Skill in most settings.
This degradation is often accompanied by an increase in average steps, which means that the agent inefficiently spends more steps on redundant exploration yet solves fewer tasks.
These results suggest that distilled skills can act as noise, impairing both task success and execution efficiency.
Verifying their reusability before admission is therefore important for filtering out non-reusable skills.

\paragraph{Effective skill verification depends on the quality of execution evidence.}
\proposed{} consistently outperforms execution-based curation methods (ExpeL, ACE, and SkillOS) across all models and benchmarks, improving SR over the strongest of these baselines by up to 21.4\%p.
These results suggest that using execution feedback alone does not guarantee effective skill verification.
ExpeL obtains evidence from repeated executions of the source task, whereas ACE and SkillOS use subsequent task stream.
\proposed{} instead acquires the verification evidence from execution on the skill-specific scenarios.
Section~\ref{sec:rq_evidence} validates the contribution of scenario synthesis by comparing with other execution-based evidence under the same rollout budget.

\paragraph{Reliable verification must account for both \textsf{Recovery} and \textsf{Regression}.}
On ALFWorld with Gemini, MemP and \proposed{} achieve the same \textsf{Recovery}, but \proposed{} has approximately 37\% lower \textsf{Regression}.
Consequently, MemP reduces SR relative to No Skill, whereas \proposed{} improves it.
Reliable verification should therefore both help the agent solve previously failed tasks and preserve success on tasks it already solves.

\paragraph{Verification selects an effective subset of the original skill library.}
Compared with the unverified Vanilla Skill library, verification increases \textsf{Recovery} by up to 9.2\%p and reduces or maintains \textsf{Regression} across all settings.
The resulting library achieves the lowest \textsf{Regression} across all settings and the highest \textsf{Recovery} in all but one setting.
It is the only library with higher \textsf{Recovery} than \textsf{Regression} in every setting, consistently improving SR over No Skill.
Together, these results show that verification selects an effective subset of the original library.

\paragraph{Gains from verification are bounded by the reusable skills in the original library.}
On ALFWorld, Gemini and Qwen(27B) already solve most held-out tasks without skills, which leaves little room for verification to improve SR, yet \proposed{} is the only library that does.
On WebShop, where the executors solve far fewer tasks on their own, the verified library adds more than 10\%p with lower \textsf{Regression}.
Verification selects from the original library rather than improving the skills themselves, so it can add only what that library contains.
Indeed, far fewer of the ALFWorld skills are helpful on held-out tasks (Section~\ref{sec:rq_validity}).

\section{Analysis}
\label{sec:analysis}
\label{sec:additional_experiments}
The main experiments show that the verified library improves downstream performance.
We now examine which components of our verification framework contribute to this improvement and whether it reflects accurate assessment of skill reusability through the following research questions:
\begingroup
\setlength{\leftmargini}{2pc}
\begin{itemize}
    \item \textbf{RQ1:} \emph{Does scenario synthesis improve the effectiveness of verification?}
    \item \textbf{RQ2:} \emph{Does verification predict skill reusability?}
    \item \textbf{RQ3:} \emph{Does reliable skill verification require a stronger model?}
    \item \textbf{RQ4:} \emph{What do the synthesized scenarios reveal about skill reusability?}
\end{itemize}
\endgroup

\Needspace{4\baselineskip}%
\subsection{RQ1: Does scenario synthesis improve the effectiveness of verification?}
\label{sec:rq_evidence}

To evaluate the contribution of synthesized scenarios, we compare verification evidence generated from synthesized scenarios, source tasks, and randomly selected tasks.
We use Gemini 3.1 Flash-Lite on both benchmarks and keep the \textsc{Verifier} and rollout budget fixed, varying only the tasks on which evidence is generated.
Under this budget, each skill is verified on 5 synthesized scenarios, 5 executions of its source task, or 5 random tasks.
We evaluate verification accuracy against each skill’s effectiveness on held-out tasks, measured through paired executions with and without the skill (Appendix~\ref{app:implementation}).
A skill is labeled helpful if it increases the success rate and harmful if it decreases.
\input{fig_rq_evidence}

Synthesized scenarios yield the highest verdict F1 and the best-performing skill library on both benchmarks (Figure~\ref{fig:rq_evidence}), with held-out SR up to 15.7\%p above the alternatives.
On ALFWorld, the verdicts from source and random tasks barely separate helpful from harmful skills, and the libraries they retain fall below No Skill, with random-task verification even below the unverified library.
Source tasks expose the skill-relevant situation only in its original context and random tasks rarely expose it at all, so neither provides the evidence that verification requires.

\par\WFclear
\Needspace{14\baselineskip}
\subsection{RQ2: Does verification predict skill reusability?}
\label{sec:rq_validity}
\label{sec:rq1}

\begin{wraptable}[12]{r}{0.42\textwidth}
\vspace{-22pt}
\centering
\footnotesize
\setlength{\tabcolsep}{2.5pt}
\renewcommand{\arraystretch}{1.0}
\caption{Verdict agreement with held-out skill effects.}
\label{tab:rq1_validity}
\vspace{4pt}
\begin{tabular}{l cccc}
\toprule
\multirow{2}{*}{\textbf{\textsc{Proposer}}} &
\multicolumn{4}{c}{\textbf{Verdict agreement (\%)}} \\
\cmidrule(lr){2-5}
& Prec. & Rec. & F1 & Spec. \\
\midrule
\rowcolor{gray!15}
\multicolumn{5}{@{}l@{}}{\textbf{WebShop}} \\
Gemini 3.1 Flash & \textbf{96.0} & \textbf{82.8} & \textbf{88.9} & \textbf{94.4} \\
GPT-5.6 Luna     & 95.2          & 69.0          & 80.0          & \textbf{94.4} \\
\midrule
\rowcolor{gray!15}
\multicolumn{5}{@{}l@{}}{\textbf{ALFWorld}} \\
Gemini 3.1 Flash & \textbf{39.1} & \textbf{81.8} & \textbf{52.9} & \textbf{60.0} \\
GPT-5.6 Luna     & 34.8          & 72.7          & 47.1          & 57.1 \\
\bottomrule
\end{tabular}
\end{wraptable}

Table~\ref{tab:rq1_validity} reports precision, recall, F1, and specificity of the verdicts on both benchmarks.
On WebShop, verification closely predicts the held-out utility of retained skills.
High precision and specificity indicate that verification reliably retains helpful skills on held-out tasks while filtering out harmful ones.
The 82.8\% recall shows that verification preserves most helpful skills, although some are incorrectly rejected.
Thus, on WebShop, a \textsc{Keep} verdict provides a strong indication that the skill remians useful beyond its source experience.

Overall agreement is lower on ALFWorld, but verification still preserves the useful skills and improves the composition of the library.
Only 23.9\% of the ALFWorld skills are helpful on held-out tasks, which lowers precision.
Verification nevertheless retains 81.8\% of the helpful skills, nearly the same recall as on WebShop, and helpful skills make up 39.1\% of the verified library, which is higher than the unverified library.
Verification thus selects a more useful subset from execution evidence even where helpful skills are scarce.

\subsection{RQ3: Does reliable skill verification require a stronger model?}
\label{sec:rq_model}
Our framework targets self-evolving agents, which the quality of verification must therefore not depend on the capability of the model.
To test this property, we replace only the \textsc{Proposer} with a stronger model (GPT-5.6 Luna~\citep{openai2026gpt56}).
The two \textsc{Proposer}s in Table~\ref{tab:rq1_validity} share the skills, the executor, the \textsc{Builder}, and the \textsc{Verifier}.
Comparing their verdicts against the same held-out effects evaluates whether a stronger model is essential for reliable verification.

On both benchmarks, the executor's own model matches or exceeds GPT-5.6 Luna in every metric and retains more of the helpful skills.
The stronger \textsc{Proposer} is not more accurate, since it rejects more helpful skills but no more harmful ones.
Reliable verdicts therefore do not depend on the intuitive judgment that only a stronger model could provide.
This result follows from the division of scenario synthesis into planning and construction, since the LLM is used only for planning and rollouts, and construction and scoring are fixed procedures.
Structured execution evidence thus lowers the dependence of verification on model capability, and a self-evolving agent can select useful skills from its own execution evidence without an external model stronger than itself.

\subsection{RQ4: What do the synthesized scenarios reveal about skill reusability?}
\label{sec:rq_types}
\definecolor{rcgold}{HTML}{D9B45B}
\definecolor{rcteal}{HTML}{6FA8B8}
\definecolor{rcsage}{HTML}{8FA895}
\definecolor{rcedge}{HTML}{2C4A66}
\definecolor{rctext}{HTML}{3B4A5C}
\pgfplotsset{rqcrit/.style={
  scale only axis, width=0.27\textwidth, height=0.15\textwidth,
  axis y line*=left, axis x line*=bottom,
  axis line style={black!75, line width=0.5pt},
  tick align=outside, major tick length=2pt, tick style={black!75, line width=0.4pt},
  tick label style={font=\scriptsize, color=black!80, /pgf/number format/assume math mode=true, inner sep=1.5pt},
  ylabel near ticks, every axis y label/.append style={font=\scriptsize, color=black!90, inner sep=0.5pt},
  title style={font=\footnotesize\bfseries, at={(axis description cs:0.5,1.04)}, anchor=base, inner sep=0pt},
  xmin=0.3, xmax=7.7, xtick={2,6}, xticklabels={ALFWorld,WebShop},
  xticklabel style={font=\scriptsize, color=black!90, inner sep=2pt},
  clip=false,
}}
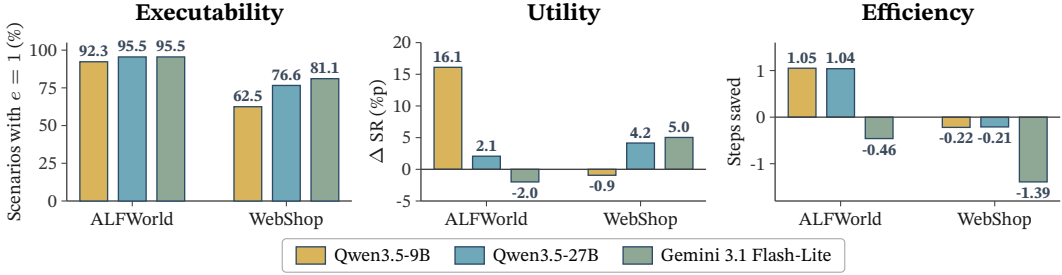
\begin{figure}[t]
\centering
{\renewcommand{\rmdefault}{stix2}\normalfont%
\begin{tikzpicture}
\begin{axis}[rqcrit, name=A, title={Executability}, ylabel={Scenarios with $e=1$ (\%)}, ymin=0, ymax=105, ytick={0,25,50,75,100}]
\path[draw=rcedge, fill=rcgold, line width=0.5pt] (axis cs:0.64,0.0) rectangle (axis cs:1.3599999999999999,92.3);
\node[anchor=south, font=\fontsize{6.2}{7}\selectfont\bfseries, text=rctext, inner sep=0.8pt, yshift=1pt] at (axis cs:1,92.3) {92.3};
\path[draw=rcedge, fill=rcteal, line width=0.5pt] (axis cs:1.6400000000000001,0.0) rectangle (axis cs:2.36,95.5);
\node[anchor=south, font=\fontsize{6.2}{7}\selectfont\bfseries, text=rctext, inner sep=0.8pt, yshift=1pt] at (axis cs:2,95.5) {95.5};
\path[draw=rcedge, fill=rcsage, line width=0.5pt] (axis cs:2.64,0.0) rectangle (axis cs:3.36,95.5);
\node[anchor=south, font=\fontsize{6.2}{7}\selectfont\bfseries, text=rctext, inner sep=0.8pt, yshift=1pt] at (axis cs:3,95.5) {95.5};
\path[draw=rcedge, fill=rcgold, line width=0.5pt] (axis cs:4.64,0.0) rectangle (axis cs:5.36,62.5);
\node[anchor=south, font=\fontsize{6.2}{7}\selectfont\bfseries, text=rctext, inner sep=0.8pt, yshift=1pt] at (axis cs:5,62.5) {62.5};
\path[draw=rcedge, fill=rcteal, line width=0.5pt] (axis cs:5.64,0.0) rectangle (axis cs:6.36,76.6);
\node[anchor=south, font=\fontsize{6.2}{7}\selectfont\bfseries, text=rctext, inner sep=0.8pt, yshift=1pt] at (axis cs:6,76.6) {76.6};
\path[draw=rcedge, fill=rcsage, line width=0.5pt] (axis cs:6.64,0.0) rectangle (axis cs:7.36,81.1);
\node[anchor=south, font=\fontsize{6.2}{7}\selectfont\bfseries, text=rctext, inner sep=0.8pt, yshift=1pt] at (axis cs:7,81.1) {81.1};
\end{axis}
\begin{axis}[rqcrit, name=B, at={($(A.east)+(26pt,0)$)}, anchor=west, title={Utility}, ylabel={$\Delta$ SR (\%p)}, ymin=-5, ymax=20, ytick={-5,0,5,10,15,20}]
\path[draw=rcedge, fill=rcgold, line width=0.5pt] (axis cs:0.64,0.0) rectangle (axis cs:1.3599999999999999,16.1);
\node[anchor=south, font=\fontsize{6.2}{7}\selectfont\bfseries, text=rctext, inner sep=0.8pt, yshift=1pt] at (axis cs:1,16.1) {16.1};
\path[draw=rcedge, fill=rcteal, line width=0.5pt] (axis cs:1.6400000000000001,0.0) rectangle (axis cs:2.36,2.07);
\node[anchor=south, font=\fontsize{6.2}{7}\selectfont\bfseries, text=rctext, inner sep=0.8pt, yshift=1pt] at (axis cs:2,2.07) {2.1};
\path[draw=rcedge, fill=rcsage, line width=0.5pt] (axis cs:2.64,0.0) rectangle (axis cs:3.36,-1.99);
\node[anchor=north, font=\fontsize{6.2}{7}\selectfont\bfseries, text=rctext, inner sep=0.8pt, yshift=-1pt] at (axis cs:3,-1.99) {-2.0};
\path[draw=rcedge, fill=rcgold, line width=0.5pt] (axis cs:4.64,0.0) rectangle (axis cs:5.36,-0.94);
\node[anchor=north, font=\fontsize{6.2}{7}\selectfont\bfseries, text=rctext, inner sep=0.8pt, yshift=-1pt] at (axis cs:5,-0.94) {-0.9};
\path[draw=rcedge, fill=rcteal, line width=0.5pt] (axis cs:5.64,0.0) rectangle (axis cs:6.36,4.15);
\node[anchor=south, font=\fontsize{6.2}{7}\selectfont\bfseries, text=rctext, inner sep=0.8pt, yshift=1pt] at (axis cs:6,4.15) {4.2};
\path[draw=rcedge, fill=rcsage, line width=0.5pt] (axis cs:6.64,0.0) rectangle (axis cs:7.36,5.03);
\node[anchor=south, font=\fontsize{6.2}{7}\selectfont\bfseries, text=rctext, inner sep=0.8pt, yshift=1pt] at (axis cs:7,5.03) {5.0};
\draw[black!75, line width=0.5pt] (axis cs:0.3,0) -- (axis cs:7.7,0);
\end{axis}
\begin{axis}[rqcrit, name=C, at={($(B.east)+(26pt,0)$)}, anchor=west, title={Efficiency}, ylabel={Steps saved}, ymin=-1.8, ymax=1.6, ytick={-1,0,1}]
\path[draw=rcedge, fill=rcgold, line width=0.5pt] (axis cs:0.64,0.0) rectangle (axis cs:1.3599999999999999,1.05);
\node[anchor=south, font=\fontsize{6.2}{7}\selectfont\bfseries, text=rctext, inner sep=0.8pt, yshift=1pt] at (axis cs:1,1.05) {1.05};
\path[draw=rcedge, fill=rcteal, line width=0.5pt] (axis cs:1.6400000000000001,0.0) rectangle (axis cs:2.36,1.04);
\node[anchor=south, font=\fontsize{6.2}{7}\selectfont\bfseries, text=rctext, inner sep=0.8pt, yshift=1pt] at (axis cs:2,1.04) {1.04};
\path[draw=rcedge, fill=rcsage, line width=0.5pt] (axis cs:2.64,0.0) rectangle (axis cs:3.36,-0.46);
\node[anchor=north, font=\fontsize{6.2}{7}\selectfont\bfseries, text=rctext, inner sep=0.8pt, yshift=-1pt] at (axis cs:3,-0.46) {-0.46};
\path[draw=rcedge, fill=rcgold, line width=0.5pt] (axis cs:4.64,0.0) rectangle (axis cs:5.36,-0.22);
\node[anchor=north, font=\fontsize{6.2}{7}\selectfont\bfseries, text=rctext, inner sep=0.8pt, yshift=-1pt] at (axis cs:5,-0.22) {-0.22};
\path[draw=rcedge, fill=rcteal, line width=0.5pt] (axis cs:5.64,0.0) rectangle (axis cs:6.36,-0.21);
\node[anchor=north, font=\fontsize{6.2}{7}\selectfont\bfseries, text=rctext, inner sep=0.8pt, yshift=-1pt] at (axis cs:6,-0.21) {-0.21};
\path[draw=rcedge, fill=rcsage, line width=0.5pt] (axis cs:6.64,0.0) rectangle (axis cs:7.36,-1.39);
\node[anchor=north, font=\fontsize{6.2}{7}\selectfont\bfseries, text=rctext, inner sep=0.8pt, yshift=-1pt] at (axis cs:7,-1.39) {-1.39};
\draw[black!75, line width=0.5pt] (axis cs:0.3,0) -- (axis cs:7.7,0);
\end{axis}
\node[draw=black!35, rounded corners=1.5pt, line width=0.4pt, inner xsep=5pt, inner ysep=2.8pt, anchor=north] at ($(A.south west)!0.5!(C.south east)+(0,-14pt)$) {%
  \raisebox{-1.3pt}{\tikz{\node[draw=rcedge, fill=rcgold, line width=0.5pt, sharp corners, inner sep=0pt, minimum width=11pt, minimum height=6.5pt] {};}}\hspace{3pt}{\scriptsize Qwen3.5-9B}\hspace{9pt}%
  \raisebox{-1.3pt}{\tikz{\node[draw=rcedge, fill=rcteal, line width=0.5pt, sharp corners, inner sep=0pt, minimum width=11pt, minimum height=6.5pt] {};}}\hspace{3pt}{\scriptsize Qwen3.5-27B}\hspace{9pt}%
  \raisebox{-1.3pt}{\tikz{\node[draw=rcedge, fill=rcsage, line width=0.5pt, sharp corners, inner sep=0pt, minimum width=11pt, minimum height=6.5pt] {};}}\hspace{3pt}{\scriptsize Gemini 3.1 Flash-Lite}};
\end{tikzpicture}}
\caption{Executability, utility, and efficiency across all executors and benchmarks.}
\label{fig:rq4_criteria}
\end{figure}
Figure~\ref{fig:rq4_criteria} analyze the skills' reusability according to three verification criterion executability, utility, and efficiency (Section~\ref{sec:verification}) across all models.
Executability is the number of scenarios with at least one action assigned $e=1$, indicating that the action implements the skill's \texttt{principle}.
Utility is the gain of the success rate between paired runs.
Efficiency compares remaining action counts from the skill-used step to termination.
In case of efficiency, we include only success pairs in both runs.

On WebShop, agents take more actions on average with skills, even when both with and without-skill runs succeed.
For Qwen3.5-27B and Gemini, skills improve task success but also add inefficiency on tasks that succeed without them.
This indicates that since success gains alone are insufficient to assess skill reusability, verification must also account for the additional actions introduced by skill guidance.
On ALFWorld, both Qwen models instead gain in success and save actions, showing that skill guidance can also yield successful execution.
Gemini exhibits a different failure pattern.
Despite high executability, skill lowers success and increases action counts, exposing a limitation in the benefit of the guidance beyond whether it can be followed.

\section{Related Work}
\label{sec:relatedwork}
\paragraph{Skill acquisition and curation.}
Agents acquire reusable skills by abstracting past executions into programs or workflows that can be retrieved and applied in subsequent tasks~\citep{wang2023voyager,wang2025awm,wang2025inducing}.
A complementary approach distills successful and failed experiences into textual insights, reasoning strategies, or procedural instructions that guide future decisions~\citep{zhao2024expel,ouyang2026reasoningbank,fang2026memp}.
Feedback-driven curation provides an implicit form of verification, using subsequent execution outcomes to revise stored guidance or learn policies for managing the skill repository~\citep{fang2026memp,zhang2026agentic,ouyang2026skillos}.
More explicit verification evaluates the effects of individual skills through source-task tests or paired runs with and without the skill, informing refinement and retention decisions~\citep{wang2025inducing,ma2026skillgen}.
However, these methods can leave skills insufficiently verified, as the verification evidence they provide are neither skill-relevant nor new.
We identify the lack of reliable opportunities to evaluate each skill as a central verification problem.

\paragraph{Task and environment synthesis.}
Recent approaches synthesize tasks and environments to train LLM agents.
Task generation expands training opportunities within available environments~\citep{zhou2025proposer,sun2025genesis,qi2025webrl,zhou2026self}.
Environment synthesis constructs the interaction setting itself through executable environments or LLM-based simulators~\citep{chae2026safe,fang2026towards,song2026envscaler,chen2026scaling}.
Adaptive approaches use the learner's performance to guide generation toward suitable difficulty levels or weaknesses in the current policy~\citep{dennis2020emergent,zala2024envgen,guo2025genenv}.
However, these objectives are insufficient for skill verification on their own, as they do not ensure that execution encounters the situation that the skill targets.
To address the challenges in skill verification, we propose a framework that dynamically synthesizes tasks and environment for each target skill at verification time.

\section{Conclusion}
\label{sec:conclusion}
We introduce \proposed{}, a framework that verifies skill reusability by dynamically synthesizing tasks and environments that reproduce each skill's applicable conditions in contexts beyond its source experience.
The framework verifies each skill by having the agent solve the synthesized tasks in the synthesized environments, testing whether its guidance remains useful in these new situations.
Across ALFWorld and WebShop, the verified libraries achieve the highest success rate and efficiency compared to other methods.
Further analyses show that synthesized scenarios improve the quality of verification, and that effective verification is possible using the executor's own model.


\section*{AI Use Statement}

We used LLMs to assist with manuscript writing and polishing for clarity and readability, as well as with implementing code for our experiments. The authors reviewed all AI-assisted text and manually verified all AI-generated code before using it in the experiments. The authors take full responsibility for the final content of this work, including all AI-assisted text and code.

\bibliography{references}
\bibliographystyle{iclr2027_conference}

\appendix
\clearpage
\addtocontents{toc}{\protect\setcounter{tocdepth}{2}}
\renewcommand{\contentsname}{Appendix Contents}
\tableofcontents
\clearpage
\section{Baseline Details}
\label{app:baselines}
We distinguish \emph{no verification} (unconditionally retaining generated skills) from \emph{execution-based verification}, which includes feedback-driven skill admission, revision, removal, or reweighting.
The latter uses either \emph{source replay} (repeated attempts at the original task) or \emph{downstream tasks} (subsequent source tasks selected independently of a skill).
All libraries are built from the same 100 source tasks and frozen before held-out evaluation.

\paragraph{Vanilla Skill.}
\emph{No verification.}
Each source trajectory is distilled into an applicability condition and an actionable principle using the skill distillation prompts in Appendix~\ref{app:prompt_distillation}.
All generated skills are retained; \proposed{} verifies skills from this same initial pool.

\paragraph{ReasoningBank.}
\emph{No verification.}
ReasoningBank sequentially solves source tasks using prior memories, judges each outcome, and extracts up to three structured memories from each trajectory.
Our basic variant, without MaTTS, appends all extracted items without revising or deleting existing ones: the judge assesses task success, not memory utility.

\paragraph{MemP.}
\emph{No verification.}
We use MemP's \emph{Script} representation with \emph{Vanilla} updates.
The executor sequentially solves source tasks using retrieved scripts; after each task, a builder appends one workflow summary regardless of success, without utility-based filtering, revision, or deletion.

\paragraph{ExpeL.}
\emph{Execution-based verification: source replay.}
ExpeL collects source trajectories before extracting insights.
After failure, our executor reflects and retries the same task at most once.
Using the same model, the extractor compares successful and failed attempts and groups of successes, curating rules through \texttt{ADD}, \texttt{EDIT}, \texttt{AGREE}, and \texttt{REMOVE} with importance counters.

\paragraph{ACE.}
\emph{Execution-based verification: downstream tasks.}
ACE solves tasks with an evolving playbook, reflects on entry helpfulness, and incrementally curates the playbook.
Our implementation makes three source passes, adding entries and updating feedback counters, with separate duplicate merging and budget-triggered pruning of harmful entries.
Held-out evaluation uses the entire frozen playbook rather than top-three retrieval.

\paragraph{SkillOS.}
\emph{Execution-based verification: downstream tasks.}
We use training-free \emph{SkillOS-base}~\citep{ouyang2026skillos}, with the paper's Appendix A prompts.
The executor makes one ordered source pass, retrieving skills from the evolving repository with BM25.
After each task, the curator uses its trajectory, outcome, and retrieved skills to insert, update, or delete skills before the next task.

\section{Implementation Details}
\label{app:implementation}
\paragraph{Executor.}
Table~\ref{tab:executor_settings} summarizes the executor settings for the two benchmarks.
At each step, the executor receives the current observation and admissible actions together with the recent interaction history.
The executor prompts are provided in Appendix~\ref{app:prompt_executor}.

\begin{table}[ht]
\centering
\small
\renewcommand{\arraystretch}{1.15}
\setlength{\tabcolsep}{10pt}
\caption{Executor settings. History length counts preceding observation--action pairs; the current observation is supplied separately.}
\label{tab:executor_settings}
\begin{tabular}{lcc}
\toprule
\textbf{Setting} & \textbf{ALFWorld} & \textbf{WebShop} \\
\midrule
Action history length & 10 & 5 \\
History content & Observation--action pairs & Observation--action pairs \\
Maximum steps per episode & 30 & 15 \\
\bottomrule
\end{tabular}
\end{table}

\paragraph{Skill distillation and retrieval.}
We distill each source trajectory into an applicability condition and an actionable principle (Appendix~\ref{app:prompt_distillation}).
On both benchmarks, held-out retrieval ranks full skill texts against the task query by cosine similarity using \texttt{text-embedding-3-small}, supplying the top three skills in descending order.

For Question~I (Section~\ref{sec:preliminary}), GPT-5.6 Luna extracts fixed conditions from \texttt{When to apply}, consulting \texttt{Principle} for context, then checks them against the task objective and with-skill trajectory.
\textsc{Observed} requires evidence for all conditions and their logical and temporal relationships; state and history conditions require trajectory evidence, not task outcomes or claimed skill use.
Otherwise, the label is \textsc{Not Observed}.
We report the mean number of \textsc{Observed} tasks per skill across 50 skills, each evaluated on 10 retrieved tasks.

For Question~II, Gemini 3.1 Flash-Lite executes a fixed random stream of 500 held-out WebShop tasks using the same 50-skill library, with three repeats per setting: Full Library, Top-3, and Top-5 Retrieval.
Retrieval uses the embedding method above.
The ReAct executor cites skill identifiers when using them; at each stream prefix, we report the percentage of skills cited in at least one or five distinct tasks.

\paragraph{\textsc{Verifier}.}
\label{app:verifier_details}
The \textsc{Verifier} compares trajectories $\tau^{+}$ and $\tau^{-}$ produced by the same executor in the same synthesized scenario, with and without skill $k$, respectively.
We align the full trajectories by maximizing the number of pairs with identical pre-action observations, preserving temporal order and using each step at most once.
Ties are resolved by choosing the lexicographically earliest sequence of paired step indices, with the $\tau^{+}$ index first.
For each distinct observation, we retain only the earliest aligned pair whose step in $\tau^{+}$ is marked as using skill $k$.
From the same observation, we examine how using the skill changes the agent's subsequent behavior compared with execution without the skill.

For each retained pair, let $t^{\pm}$ denote the matched step indices and $T^{\pm}$ the termination indices of the corresponding trajectories.
The remaining action counts $T^{\pm}-t^{\pm}$ include all subsequent actions in both trajectories, including those taken when the same observation recurs.
To assess utility, we compare terminal rewards $r^{\pm}\in\{0,1\}$, which are $1$ for task success and $0$ for failure under the scenario's success criterion.
To account for efficiency, we discount these rewards back to the matched steps so that successful continuations requiring fewer actions receive higher returns:
\begin{equation}
\Delta u
=r^{+}\gamma^{T^{+}-t^{+}}
-r^{-}\gamma^{T^{-}-t^{-}},
\qquad 0<\gamma<1.
\label{eq:utility_gain}
\end{equation}

Executability concerns whether the \texttt{principle} can be executed as admissible actions supported by the environment.
The \textsc{Verifier} uses agreement between the \texttt{principle} and an action in $\tau^{+}$ as evidence of this feasibility at each matched point.
It sets $e=1$ when the action implements the corresponding instruction and $e=0$ when this evidence is absent.
Only points with observed evidence of executability contribute to $R(k)$.
We first average $e\,\Delta u$ over the retained pairs within each trajectory pair, then across synthesized scenarios.
This empirical averaging gives the reusability score and \textsc{Keep}/\textsc{Reject} verdict in Equation~\ref{eq:skill_retention}.

\paragraph{Evaluation protocol.}
For RQ1 (Section~\ref{sec:rq_validity}), we execute the 500 held-out WebShop tasks with and without each of the 50 evaluated skills.
Let $\mathrm{Help}_k$ denote the number of held-out tasks that skill $k$ turns from failure to success and $\mathrm{Harm}_k$ the number it turns from success to failure; the effectiveness of the skill is calculated $E_k=(\mathrm{Help}_k-\mathrm{Harm}_k)/500$.
Spearman's $\rho$, with average ranks assigned to ties, and Pearson's $r$ are computed between $E_k$ and the reusability score $R(k)$ assigned by the verification pipeline over all 50 skills.
For the agreement metrics, a skill is labeled helpful when $\mathrm{Help}_k>\mathrm{Harm}_k$ and harmful when $\mathrm{Help}_k<\mathrm{Harm}_k$.

\section{Scenario Synthesis}
\label{app:scenario_synthesis}

\paragraph{Synthesis Types}
\label{app:typology}
The synthesis typology of Section~\ref{sec:rq_types} covers scenarios generated for WebShop and ALFWorld skills in the Qwen3.5-27B setting.
For each environment, GPT-5.6 Luna first summarizes each scenario's operations and implementation plan in one or two sentences and induces types from these summaries.
To categorize how skills cause harm, we select the scenarios, in which the executor fails with the skill, succeeds without it, and receives a negative scenario score.
Given the skill, both trajectories, and their outcomes, the same model explains how the skill could have contributed to the failure, and the explanations are grouped into types in the same way.
\definecolor{rqtfill}{HTML}{6FA8B8}
\definecolor{rqtedge}{HTML}{2C4A66}
\definecolor{rqttext}{HTML}{3B4A5C}
\pgfplotsset{rqtypes/.style={
  scale only axis, width=0.245\textwidth, height=112.0pt, y dir=reverse,
  xmin=0, xmax=100, xtick={0,25,50,75,100}, xlabel={Share of scenarios (\%)},
  axis y line*=left, axis x line*=bottom,
  axis line style={black!75, line width=0.5pt},
  tick align=outside, major tick length=2pt, tick style={black!75, line width=0.4pt},
  tick label style={font=\scriptsize, color=black!80, /pgf/number format/assume math mode=true, inner sep=1.5pt},
  yticklabel style={font=\scriptsize, color=black!90, inner sep=0pt, xshift=-4pt, text width=78pt, align=right},
  ytick style={draw=none},
  every axis x label/.append style={font=\scriptsize, color=black!90, at={(axis description cs:0.5,0)}, anchor=north, yshift=-8pt},
  title style={font=\footnotesize\bfseries, color=rqttext, at={(axis description cs:0.171,1)}, anchor=base, yshift=7pt, inner sep=0pt},
  nodes near coords={\pgfmathprintnumber[fixed, fixed zerofill, precision=1]{\pgfplotspointmeta}},
  point meta=x,
  every node near coord/.append style={font=\fontsize{6.5}{7.3}\selectfont\bfseries, color=rqttext, anchor=west, xshift=2.5pt, inner sep=0pt, /pgf/number format/assume math mode=true},
  clip=false,
}}
\begin{figure}[!htb]
\centering
{\renewcommand{\rmdefault}{stix2}\normalfont%
\begin{tikzpicture}
\begin{axis}[rqtypes, name=A, title={(a) WebShop},
  ymin=0.5, ymax=8.5, ytick={1,2,3,4,5,6,7,8}, yticklabels={{Add similar distractors},{Rank target higher},{Edit target attributes},{Rank target lower},{Change target options},{No environment change},{Edit other candidates},{Move or reword info}}]
\addplot[xbar, bar width=9.0pt, fill=rqtfill, draw=rqtedge, line width=0.5pt] coordinates {(79.6,1) (74.7,2) (21.4,3) (13.2,4) (9.9,5) (7.9,6) (7.2,7) (3.9,8)};
\end{axis}
\begin{axis}[rqtypes, name=B, at={($(A.south east)+(102pt,0)$)}, anchor=south west, title={(b) ALFWorld},
  ymin=0.5, ymax=7.5, ytick={1,2,3,4,5,6,7}, yticklabels={{Constrain object access},{Require state changes},{Add search distractors},{Increase target count},{Seed interaction history},{Set transfer layout only},{Set inspection cues}}]
\addplot[xbar, bar width=9.0pt, fill=rqtfill, draw=rqtedge, line width=0.5pt] coordinates {(43.2,1) (42.5,2) (16.2,3) (15.8,4) (9.5,5) (9.3,6) (6.4,7)};
\end{axis}
\end{tikzpicture}}
\caption{Planned synthesis mechanisms for Qwen3.5-27B on (a) WebShop and
(b) ALFWorld.
Each bar shows the percentage of scenarios using that transformation.
A scenario may involve multiple transformations.}
\label{fig:rq4_types}
\end{figure}
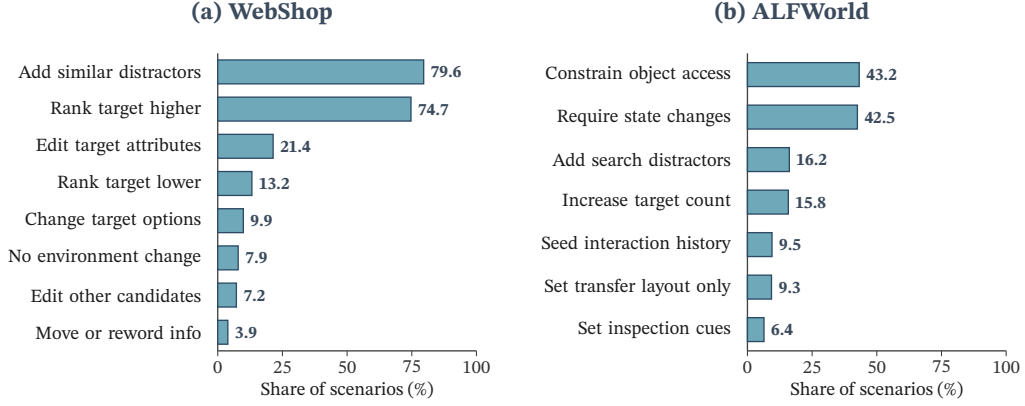

Figure~\ref{fig:rq4_types} reports the share of scenarios in each synthesis type; types are not exclusive, so shares sum to more than 100\%.

\section{Case Study}
\label{app:case_study}

\definecolor{csred}{HTML}{C4161C}
\definecolor{csgreen}{HTML}{2E7D32}
\definecolor{csgrey}{HTML}{6B7280}
\definecolor{csshade}{HTML}{EAF1F8}%
\providecommand{\csact}[1]{{\fontsize{7}{8}\selectfont\fontfamily{cmtt}\selectfont #1}}
\providecommand{\csstep}[1]{\makebox[9pt][r]{\textcolor{csgrey}{\scriptsize #1}}}
\providecommand{\csobs}[1]{{\color{csgrey}$\rightarrow$~#1}}
\providecommand{\csnote}[1]{\textcolor{csgrey}{#1}}
\providecommand{\cshl}[1]{{\setlength{\fboxsep}{0.5pt}\colorbox{csred!12}{#1}}}
\providecommand{\csbad}[1]{\textcolor{csred}{\textbf{#1}}}
\providecommand{\csgood}[1]{\textcolor{csgreen}{\textbf{#1}}}
\providecommand{\csok}{\textcolor{csgreen}{\ding{51}}}
\providecommand{\csitem}{\par\noindent\hangindent=6pt\hangafter=1\textbullet\hspace{2.5pt}}
\providecommand{\cshead}[1]{\textbf{#1}\par\nobreak}
\providecommand{\csskill}[1]{{\setlength{\fboxsep}{3pt}\noindent\colorbox{csshade}{\parbox[t]{\dimexpr\linewidth-2\fboxsep\relax}{\raggedright #1}}}}
\providecommand{\cssteps}{\begin{tabular}[t]{@{}r@{\hspace{2pt}}l@{}}}
\providecommand{\csstepsend}{\end{tabular}}
\providecommand{\cspair}[2]{\begin{tabular}[t]{@{}>{\raggedright\arraybackslash}p{111pt}@{\hspace{4pt}}>{\raggedright\arraybackslash}p{111pt}@{}}#1&#2\end{tabular}}
\providecommand{\csgroup}[2]{\rowcolor{gray!15}[0pt][0pt]\multicolumn{2}{@{}l@{}}{\textbf{#1}\quad Verdict: \csbad{\textsc{#2}}}\\}
\providecommand{\csskillrow}[3]{\multicolumn{2}{@{}>{\raggedright\arraybackslash}p{\textwidth}@{}}{\textbf{Task}\enspace\emph{#1}\par\smallskip\csskill{\texttt{condition}:\ #2\par\texttt{principle}:\ #3}}\\}
\providecommand{\csscen}[1]{\multicolumn{1}{@{}>{\raggedright\arraybackslash}p{164.5pt}|}{#1}}
\providecommand{\cscolhead}{\midrule\csscen{\textbf{Synthesized Scenario}} & \textbf{Execution Result}\\\midrule}
\begin{table}[!htb]
\centering
\caption{Case study of rejected skills. The shaded block is the skill under verification, and the red bold text marks the part of its \texttt{principle} that causes the rejection. $\tau^{+}$ and $\tau^{-}$ denote the executions of the same synthesized scenario with and without the skill, and the highlighted action is the one guided by the skill.}
\label{tab:case_study}
\fontsize{8}{9.6}\selectfont
\setlength{\tabcolsep}{3pt}
\renewcommand{\arraystretch}{1.08}
\begin{tabular}{@{}>{\raggedright\arraybackslash}p{164.5pt}>{\raggedright\arraybackslash}p{226pt}@{}}
\toprule
\csgroup{ALFWorld}{Reject}
\midrule
\csskillrow{Examine the tissuebox with the desklamp}%
  {The task requires using a stationary appliance on a movable object located in a different area.}%
  {\csbad{Locate and activate the stationary tool first}, retrieve the target and transport it to the tool's location.}
\cscolhead
\csscen{%
\cshead{Task}
\emph{put a cool bread in countertop}
\par\medskip
\cshead{Environment}
\csitem bread on the diningtable, \textbf{not cooled}
\csitem fridge in a \textbf{different area}, closed
\csitem agent starts at the diningtable; the bread can be picked up immediately}
&
\cspair{%
\csbad{$\tau^{+}$: Fail, 30 steps}
\par\smallskip
\cssteps
\csstep{1} & \cshl{\csact{go to fridge}}\\
\csstep{2} & \csact{open fridge}\\
\csstep{3} & \csact{close fridge}\\
\csstep{4} & \csact{go to cabinet}\\
\csstep{$\cdots$} & \csnote{searches other receptacles;}\\
            & \csnote{never returns for the bread}\\
\csstepsend
}{%
\csgood{$\tau^{-}$: Success, 6 steps}
\par\smallskip
\cssteps
\csstep{1} & \csact{take bread from diningtable}\\
\csstep{2} & \csact{go to fridge}\\
\csstep{3} & \csact{open fridge}\\
\csstep{4} & \csact{cool bread with fridge}\,\csok\\
\csstep{5} & \csact{go to countertop}\\
\csstep{6} & \csact{put bread on countertop}\,\csok\\
\csstepsend
}
\\
\midrule
\csgroup{WebShop}{Reject}
\midrule
\csskillrow{I need some hair drying towels that are easy to clean, and price lower than 50.00 dollars}%
  {Evaluate a product for an easy-to-clean requirement after confirming its price and category criteria.}%
  {Verify the claim in \csbad{both Features and Description} before purchase.}
\cscolhead
\csscen{%
\cshead{Task}
\emph{I want to purchase a hairdressing backpack with an easy clean feature for under 50.}
\par\medskip
\cshead{Environment}
\csitem required attribute \csact{easy clean}, price at most \$50, product kind hairdressing backpack
\csitem search query \csact{hairdressing backpacks easy to clean}; the episode starts on the results page
\csitem target product placed at \textbf{rank 3}}
&
\cspair{%
\csbad{$\tau^{+}$: Fail, 15 steps}
\par\smallskip
\cssteps
\csstep{$\cdots$} & \\
\csstep{4} & \cshl{\csact{click[Description]}}\\
\csstep{5} & \csact{click[Back to Search]}\\
\csstep{6} & \csact{search[same query]}\\
\csstep{$\cdots$} & \csnote{7--15: reopens the product,}\\
            & \csnote{alternates the two sections;}\\
            & \csnote{no purchase}\\
\csstepsend
}{%
\csgood{$\tau^{-}$: Success, 4 steps}
\par\smallskip
\cssteps
\csstep{$\cdots$} & \\
\csstep{4} & \csact{click[Buy Now]}\,\csok\\
\csstepsend
}
\\
\bottomrule
\end{tabular}
\end{table}

Table~\ref{tab:case_study} illustrates how synthesized scenarios expose the skill-relevant situation and use it as verification evidence.

\paragraph{ALFWorld.}
In the source task, the agent successfully examines a tissue box with a desk lamp by activating the lamp before retrieving the tissue box from another location.
This ordering works because the lamp can be activated independently and remains on while the agent retrieves the object.
The distilled skill generalizes this source-specific ordering into a broader principle: activate a stationary appliance before retrieving the target object.
To reproduce the skill's condition, the synthesized scenario places a stationary appliance (fridge) and a movable object (bread) at different locations.
The Builder places the uncool bread at the agent's starting location and keeps the fridge closed elsewhere, requiring the agent to cool the bread before placing it on a countertop.
Unlike the lamp, cooling requires the agent to hold the bread at the fridge.
Nevertheless, $\tau^{+}$ follows the skill and goes to the fridge first, then fails to return for the bread and reaches the step limit.
In contrast, $\tau^{-}$ takes the bread first, carries it to the fridge, cools it, and completes the task in six steps.
The scenario thus shows that the principle distilled from the source does not transfer to this new context, leading the verifier to reject the skill.

\paragraph{WebShop.}
The source task asks for an easy-to-clean hair-drying towel under \$50.
During the source experience, the agent finds a candidate whose cleaning information appears in both the Features and Description sections, motivating a skill that requires checking both before purchase.
For verification, the synthesized task instead asks for an easy-to-clean hairdressing backpack under \$50 and places a qualifying product at rank 3 of the search results.
This creates a new instance of the same product-verification situation, but one in which sufficient evidence is already available from a single section.
After inspecting the Features, $\tau^{-}$ treats the requirement as satisfied and purchases the product in four steps.
$\tau^{+}$, however, follows the principle and continues to seek confirmation from the Description.
Because the same claim is absent there, it repeatedly revisits the product sections and exhausts the action budget without purchasing.
The synthesized scenario therefore exposes that requiring evidence from both sections is not generally necessary, but reflects the information layout encountered in the source task, leading the verifier to reject the skill.

\newtcolorbox{promptbox}[1]{%
  enhanced,
  breakable,
  colback=black!1,
  colframe=black!50,
  colbacktitle=black!65,
  coltitle=white,
  fonttitle=\normalfont\rmfamily\bfseries,
  fontupper=\normalfont\rmfamily,
  title={#1},
  title after break={#1\ (continued)},
  titlerule=0pt,
  boxrule=0.5pt,
  arc=3pt,
  boxsep=0pt,
  left=10pt,
  right=10pt,
  top=7pt,
  bottom=7pt,
  toptitle=4pt,
  bottomtitle=4pt,
  before skip=10pt,
  after skip=10pt,
  before upper={\raggedright\setlength{\parindent}{0pt}\setlength{\parskip}{6pt}},
}

\section{Additional Results}
\label{app:results}
\paragraph{WebShop score correlation.}
Table~\ref{tab:webshop_score_correlation} supplements RQ1 with the correlations between the verification score $R(k)$ and the net held-out effect $E_k$ on WebShop, using the evaluation protocol in Appendix~\ref{app:implementation}.
Both \textsc{Proposer}s have positive Spearman correlations, indicating that skills with higher verification scores tend to have larger held-out effects.
Pearson's $r$ provides a supplementary check of a linear relation; it does not establish that the score is calibrated to the magnitude of the improvement.
\begin{table}[htbp]
\centering
\small
\caption{WebShop score correlation with held-out skill effects. $\rho$ and $r$ denote Spearman and Pearson correlations.}
\label{tab:webshop_score_correlation}
\begin{tabular}{lcc}
\toprule
\textbf{\textsc{Proposer}} & \textbf{Spearman $\rho$} & \textbf{Pearson $r$} \\
\midrule
Gemini 3.1 Flash & \textbf{0.665} & 0.552 \\
GPT-5.6 Luna & 0.662 & \textbf{0.645} \\
\bottomrule
\end{tabular}
\end{table}

\paragraph{Retrieval size.}
Table~\ref{tab:retrieval_k} examines how the number of retrieved skills affects downstream task success.
We vary the retrieval size $k\in\{1,3,5\}$ with Gemini 3.1 Flash-Lite as the executor, while keeping the \textsc{Keep}/\textsc{Reject} verdicts, the retained skill pool, the held-out tasks, and the executor settings fixed.
For each task, we rank the retained skills by cosine similarity between embeddings of the task instruction and the full skill text, using the same embedding model as in the main experiments.
The top-$k$ skills are injected into the executor's prompt.
We set $k=3$ in all other experiments because it achieves the best SR on both benchmarks, tying with $k=5$ on ALFWorld and outperforming both $k=1$ and $k=5$ on WebShop.
\begin{table}[h]
\centering
\small
\renewcommand{\arraystretch}{1.1}
\setlength{\tabcolsep}{8pt}
\caption{Held-out SR (\%) of Gemini 3.1 Flash-Lite under different retrieval sizes $k$, with the number of solved tasks in parentheses. $k=3$ is the setting of all other experiments. Bold marks the best value in each row.}
\label{tab:retrieval_k}
\begin{tabular}{lccc}
\toprule
\textbf{Benchmark} & $k=1$ & $k=3$ & $k=5$ \\
\midrule
ALFWorld (140 tasks) & 58.6 (82) & \textbf{60.0} (84) & \textbf{60.0} (84) \\
WebShop (500 tasks)  & 31.4 (157) & \textbf{36.2} (181) & 35.0 (175) \\
\bottomrule
\end{tabular}
\end{table}

\paragraph{Causes of rejection.}
Figure~\ref{fig:rejected_causes_full} shows the primary failure causes of the rejected skills according to the verification criterion for each executor.
A rejected skill can fail several criteria; within each criterion, one primary cause is assigned per skill, and the three largest cause types are shown with the rest pooled as Other.
\input{fig_rejected_causes_full}

\clearpage
\section{Prompts}
\label{app:prompts}

We present prompt templates for execution and skill distillation, and structured summaries of the scenario-proposal prompts.
Braced placeholders denote runtime inputs, such as a task, trajectory, skill, or environment interface.
The proposal summaries describe the inputs and output schemas for each stage; retry and format-repair messages are omitted.

\subsection{Executor}
\label{app:prompt_executor}
The following prompts execute tasks without skills, using the current task, observation, admissible actions, and recent action history.

\begin{promptbox}{Executor: ALFWorld}
Follow the user's ALFRED task instructions.\\
Return only JSON with fields in this order: \texttt{\{"rationale":\allowbreak{} "step-by-step reasoning",\allowbreak{} "action":\allowbreak{} "one exact admissible action"\}}.

You are an expert agent operating in the ALFRED Embodied Environment. Your task is to: \texttt{\{task\}}\\
Below are the most recent \texttt{\{history\_\allowbreak{}length\}} observations and the corresponding actions you took: \texttt{\{action\_\allowbreak{}history\}}\\
Your current observation is: \texttt{\{current\_\allowbreak{}observation\}}\\
Your admissible actions of the current situation are: \texttt{\{admissible\_\allowbreak{}actions\}}.

Now it's your turn to take an action. You should first reason step-by-step about the current situation. Once you've finished your reasoning, choose one admissible action for the current step.\\
Return only JSON: \texttt{\{"rationale":\allowbreak{} "step-by-step reasoning",\allowbreak{} "action":\allowbreak{} "one exact admissible action"\}}.
\end{promptbox}

\begin{promptbox}{Executor: WebShop}
You are an expert autonomous agent operating in the WebShop e-commerce environment. Your task is to: \texttt{\{task\_\allowbreak{}description\}}.\\
Below are your most recent \texttt{\{history\_\allowbreak{}length\}} interaction steps (observations when provided, and actions):\\
\texttt{\{action\_\allowbreak{}history\}}

You are now at step \texttt{\{current\_\allowbreak{}step\}} out of \texttt{\{total\_\allowbreak{}steps\}} steps, and your current observation is: \texttt{\{current\_\allowbreak{}observation\}}

Your admissible actions of the current situation are: [\\
\texttt{\{available\_\allowbreak{}actions\}} ].

Now it’s your turn to take one action for the current step. You should first reason step-by-step about the current situation, then think carefully which admissible action best advances the shopping goal.

- Thought: Reasoning process\\
- Action: Return exactly one executable admissible command for current step

Return only a valid JSON object with exactly these two fields, using \texttt{rationale} for Thought and \texttt{action} for Action: \texttt{\{"rationale":\allowbreak{} "the complete Thought described above",\allowbreak{} "action":\allowbreak{} "the single executable Action described above"\}}
\end{promptbox}

\Needspace*{20\baselineskip}
\subsection{Skill distillation}
\label{app:prompt_distillation}
These prompts generate one skill from the source trajectory, before verification or revision.
The trajectory input includes the source task, observed transitions, recorded actions, and execution outcome.
The output consists of a \texttt{When to apply} condition and a \texttt{Principle} in one or two sentences.
We distinguish prompts for successful and failed source trajectories.

\begin{promptbox}{Skill distillation: ALFWorld failure}
You are an expert at extracting reusable skills from agent failures. Analyze the failed trajectory from an AI agent operating in the ALFRED embodied household environment (ALFWorld).

Trajectory: \texttt{\{trajectory\}}

Generate one concise skill that captures the reusable lesson from this failure.

- When to apply: Specific trigger condition.\\
- Principle: The core actionable insight in 1-2 sentences.
\end{promptbox}

\begin{promptbox}{Skill distillation: ALFWorld success}
You are an expert at extracting reusable skills from agent successes. Analyze the successful trajectory from an AI agent operating in the ALFRED embodied household environment (ALFWorld).

Trajectory: \texttt{\{trajectory\}}

Generate one concise skill that captures the reusable lesson from this success.

- When to apply: Specific trigger condition.\\
- Principle: The core actionable insight in 1-2 sentences.
\end{promptbox}

\begin{promptbox}{Skill distillation: WebShop failure}
You are an expert at extracting reusable skills from agent failures. Analyze the failed trajectory from an AI agent operating in an online shopping environment (WebShop).

Trajectory: \texttt{\{Trajectory\}}

Generate one concise skill that captures the reusable lesson from this failure.

- When to apply: Specific trigger condition.\\
- Principle: The core actionable insight in 1-2 sentences.
\end{promptbox}

\begin{promptbox}{Skill distillation: WebShop success}
You are an expert at extracting reusable skills from agent successes. Analyze the successful trajectory from an AI agent operating in an online shopping environment (WebShop).

Trajectory: \texttt{\{Trajectory\}}

Generate one concise skill that captures the reusable lesson from this success.

- When to apply: Specific trigger condition.\\
- Principle: The core actionable insight in 1-2 sentences.
\end{promptbox}

\subsection{\textsc{Proposer}}
\label{app:prompt_proposer}
For brevity, we summarize the proposal prompts by their inputs, core instructions, and required JSON outputs.
Each summary combines condition interpretation and scenario specification for one environment.

\begin{promptbox}{Proposer: ALFWorld}
\textbf{Input}

\texttt{\textless{}skill\_condition\textgreater{}}\\
\texttt{\textless{}source\_task\textgreater{}}\\
\texttt{\textless{}source\_trajectory\_summary\textgreater{}}\\
\texttt{\textless{}environment\_interface\textgreater{}}\\
\texttt{\textless{}skill\textgreater{}}\\
\texttt{\textless{}interpreted\_requirements\textgreater{}}\\
\texttt{\textless{}compatible\_or\_fixed\_goals\textgreater{}}\\
\texttt{\textless{}earlier\_scenario\_summaries\textgreater{}}\\
\texttt{\textless{}selected\_test\_specification\textgreater{}} (optional)

\textbf{Core instructions}

1. Preserve the full condition, including alternatives, negation, quantities, and prior events. Separate task objectives from the current state and execution history; do not derive applicability requirements from the principle.

2. Determine whether the condition is established by the initial state, real setup history, or the executor's own rollout. Leave exploration and manipulation to the executor when the condition arises during execution.

3. Specify symbolic roles and supported construction operations. Vary objects, placements, and compatible task context while preserving required capabilities and any fixed goal; concrete entities are bound later.

4. Keep the goal and the skill-guided decision unfinished. Do not invent observations or capabilities, pre-execute the principle, or design the scenario to guarantee a favorable verdict.

\textbf{Output}

\emph{Condition interpretation:} A JSON object with \texttt{status}, \texttt{encounter\_mode}, \texttt{encounter\_reason}, \texttt{history\_required}, and \texttt{task\_checks}.

\emph{Scenario specification:} One symbolic program with \texttt{roles}, \texttt{operations}, \texttt{prepare\_start}, \texttt{setup\_events}, \texttt{condition\_checks}, \texttt{test\_checks}, and \texttt{novelty}; include \texttt{goals} only when no fixed goal is supplied. Condition checks concern the encounter, while test checks concern the initial configuration.
\end{promptbox}

\begin{promptbox}{Proposer: WebShop}
\textbf{Input}

\texttt{\textless{}skill\_condition\textgreater{}}\\
\texttt{\textless{}environment\_interface\textgreater{}}\\
\texttt{\textless{}skill\textgreater{}}\\
\texttt{\textless{}interpreted\_requirements\textgreater{}}\\
\texttt{\textless{}fixed\_target\_product\textgreater{}}\\
\texttt{\textless{}linked\_human\_task\textgreater{}}\\
\texttt{\textless{}scoring\_conditions\textgreater{}}\\
\texttt{\textless{}earlier\_scenario\_summaries\textgreater{}}\\
\texttt{\textless{}response\_schema\textgreater{}}\\
\texttt{\textless{}frozen\_design\textgreater{}}\\
\texttt{\textless{}operator\_definitions\textgreater{}}

\textbf{Core instructions}

1. Restate every condition clause without adding restrictions from the principle. Preserve alternatives, history, and local absence claims; report a missing native capability rather than weaken the condition.

2. Keep the fixed target's identity and product type. Preserve the linked task and scoring conditions except for changes needed to reproduce the condition; use existing product facts whenever sufficient.

3. Specify minimal supported edits to product records, options, or result lists. Start at the native search screen with no preselected options, and use one search query derived from the final task. Leave inspections and selections to the executor.

4. Translate the frozen design into symbolic operations and identify which requirements they establish. Do not fabricate observed evidence, pre-execute the principle, or guarantee that the skill helps; an empty operation list is valid when no edits are needed.

\textbf{Output}

\emph{Condition interpretation:} A JSON object containing \texttt{status}, scoped \texttt{requirements} (task, catalog, or trajectory), and \texttt{rationale}; if unsupported, return \texttt{status} and \texttt{reason}.

\emph{Scenario design:} A JSON object with \texttt{mode}, \texttt{intervention}, \texttt{goal}, \texttt{wording}, and \texttt{novelty}. Goal fields express requirements for subsequent binding.

\emph{Operation specification:} A JSON object with \texttt{value\_declarations}, \texttt{operations}, and \texttt{coverage}, linking requirements to edits and the intended evidence locations.
\end{promptbox}

\end{document}